\documentclass[11pt]{article}
\usepackage{acl2015}
\usepackage{times}
\usepackage{url}
\usepackage{latexsym}
\usepackage{graphicx}
\usepackage{booktabs}
\usepackage{multirow}

\title{An LSTM Recurrent Network for Step Counting}

\author{Ziyi Chen \\
  Computer Science  \\
  Unversity of California Santa Cruz\\
  {\tt zchen139@ucsc.edu}}

\date{}

\begin{document}
\onecolumn

{
    {\setcounter{page}{0}\renewcommand{\thepage}{\roman{page}}}
  
  \enlargethispage{\footskip}

\maketitle
\begin{abstract}

Smartphones with sensors such as accelerometer and gyroscope can be used as pedometers and navigators. In this paper, we propose to use an LSTM recurrent network for counting the number of steps taken by both blind and sighted users, based on an annotated smartphone sensor dataset, WeAllWork. The models were trained separately for sighted people, blind people with a long cane or a guide dog for Leave-One-Out training modality. It achieved 5\% overcount and undercount rate.

\end{abstract}

\section{Introduction}

With the increasing ubiquity of smartphones, users are now carrying plenty of sensors with them such as accelerometer, gyroscope, magnetometer, wherever they go. Step counters are being integrated into an increasing number of portable consumer electronic devices such as music players, smartphones, and mobile phones \cite{wiki:pedometer}. There are various of step counting apps on smartphones. Step counters can also be used for estimating the distance and the position in indoor pedestrian navigation systems, which is especially helpful not only for blind people, but also for sighted people who need directional information in unfamiliar places.

In this paper, we propose an LSTM model trained with indoor walking sensor data from iPhones, and annotated data generated by a shoes-mounted sensor (MetaWear-CPRO) from the WeAllWalk dataset, to predict the number of steps. The MetaWear-CPRO units are attached to the uesr's shoes so they can provide precise heel strike times. In the dataset, blind volunteers using a long cane or a guide dog and sighted volunteers have different gait patterns, so we built the models separately for each group of walkers and measured the performances based on three error metrics. We also tested the accuracy of those models by using leave-one-out cross-validation. Three error metrics splitting intervals differently were used for measuring the overcount and undercount rates of the system. We tried different parameters for the LSTM models such as the number of training steps and window size to find the best settings. The model achieved 1\% overcount and undercount rate on a mixed dataset and 5\% under the Stratified Leave-One-Out training modality.

\section{Background and Related Work}

\subsection{Step Counting}

\begin{figure*}[htb]
\centering
\includegraphics[scale=0.5]{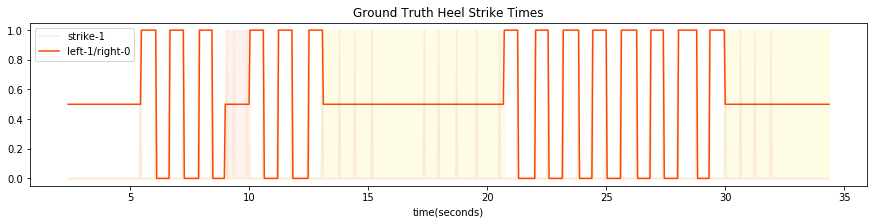}
\caption{An example of heel strike times in T1\_ID1\_WC.xml from 2.4s to 34.4s. The first step at 5.4761s is a left step. A feature motion (the participant walked into the wall, stopped, and moved to the left) happened during 9.041862s to 10.041862s. Then one turn from east to south and another turn from south to east occured during 13.1752s to 20.7419, and 30.0419s to 39.1419s respectively. Feature data, turn data, and all data recorded before the first step are removed.}
\label{fig:ground_truth}
\end{figure*}

Automatic step counting has received substantial attention in both research and commercial domains. There is a wealth of studies on the use of inertial sensors for detecting and characterizing walk-related activities. Pedometers are usually portable and electronic or electromechanical \cite{wiki:pedometer}. They can be embedded in shoes, in smartwatchs, in  smartphones, and attached to users' ankles or hung on the belt.

\cite{brajdic2013walk} evaluated common walk detection (WD) and step counting (SC) algorithms applied to smartphone sensor data. The results favor the use of standard deviation thresholding (WD) and windowed peak detection (SC) with error rates of less than 3\%.
\cite{tomlein2012advanced} introduced step detection and intelligent detection of cheating based on smartphone sensors.
\cite{naqvib2012step} presented a method for counting the number of steps taken by a user using the smartphone-based accelerometer while walking at any variable speed.

With the rapid deveolpment of deep learning, this advanced technology is used in various fields including step counting. Since sensors provide time series data, researchers have tried to use RNN for counting the number of steps. \cite{edel2015advanced} uses Bidirectional Long Short-Term Memory Recurrent Neural Networks (BLSTM-RNNs) for step detection, step length approximation as well as heading estimation.

In addition to sighted people, some researchers also pay attention to people with visual impairment. \cite{haegele2015validation} validated Centrios talking pedometer for adolescents with different level of visual impairment under daily-living condition, while  \cite{holbrook2011validation} validated the same talking pedometer for adults.

\subsection{The WeAllWalk Dataset}

The WeAllWalk dataset \cite{flores2016weallwalk} contains sensor data gathered from ten blind participants with a long cane or a guide dog and five sighted participants. The smartphone (iPhone 6s) carried by participants recorded the sensor data. Two small sensors attached to the participants’ shoes recorded heel strike times as the labels. Each segment of paths is marked as either a straight segment or a turn segment. Special circumstances on the road are also labeled as features. \cite{flores2016weallwalk} have tested six different algorithms for step counting and turn detection.

\section{Implementation}
We propose an LSTM model trained with indoor walking sensor data from iPhone for step counting. Since blind participants using a long cane or a guide dog and sighted volunteers have different gait patterns, we build the models separately and calculate the error rates by using three metrics. Also, we only consider straight segments in the paths, since the gait pattern is more likely to be regular than the gait pattern of turn segments, and remove segments marked as features such as opening a closed door and avoiding an obstacle. We try different parameters for the LSTM model and add a dropout layer to make the result more robust.

\subsection{Data Preprocessing}

\begin{figure*}[ht]
\centering
\includegraphics[scale=1]{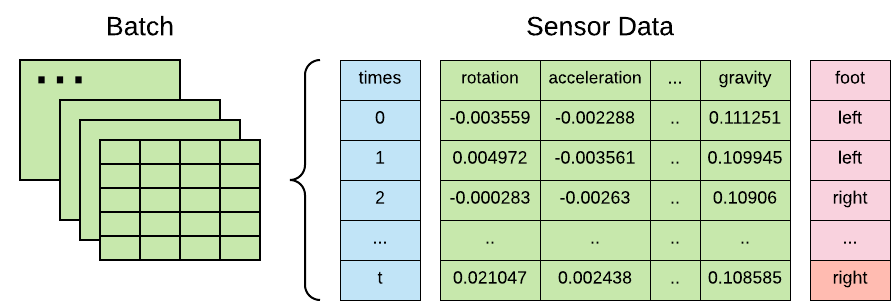}
\caption{Input (green) and output (pink) examples for LSTM. Use recorded sensor data from t previous timesteps ago to current time (green blocks, containing t records) to predict the current step (last pink block, last output of the lstm model). The intermediate outputs (light pink) of the model would be regards as output for first (t-1) records at beginning of each segment since they don't have enough previous record. }
\label{fig:batch_sensor_data}
\end{figure*}

We use two kinds of file from the dataset. The first type is the CSV files of iPhone sensor data.  Each CSV file contians recorded one iPhone sensor data per path and per person. There are 39 columns of sensor data and some of them are more useful than others such as rotation rate and user acceleration. So we use the six sensor data (rotationRateX, rotationRateY, rotationRateZ,  userAccelerationX, userAccelerationY, userAccelerationZ) as input. 

The second type is the XML files containing annotated ground truth data for all the paths walked by all the participants. Each file contains start time, end times and direction for each segment the user walked through . For each segment, the time and foot of each step is recorded, and special situation liking moving to the wall is also marked as feature with start time, end time and detailed events. We only train and test the step counting system on sensor data when participants traverse straight segments in the paths, where gait patterns is assumed to be regular, and remove time slots marked as feature. 

Since the XML files only contain heel strike times, we cannot use it as labels directly. If we regard heel strike times as 1 and other times as 0, then the labels are unbalanced. So we transform the XML file to binary signal in which left steps trigger the signal from 0 to 1 and right steps trigger the signal from 1 to 0 as shown in picture \ref{fig:ground_truth}.

Finally we windowed the data (window size is equal to timesteps) for corresponding sensor data and ground truth signals to generate the training data and labels.

\subsection{LSTM Model}
Long short-term memory (LSTM) network is a time-recursive neural network, suitable for processing and prediction the important events of time series in the relatively long delay . LSTM has many applications in science and technology domains. The LSTM-based system can learn tasks such as translating languages, image analysis, speech recognition, controlling chat robots and so on.

Long short-term memory (LSTM) network is a special kind of recurrent neural network (RNN). The main difference between LSTM and RNN lies in that it adds a "processor" to judge whether the information is useful or not in the algorithm. The structure of this processor is called cell.
Three multiplicative units (input door, forgotten door and output door) are placed in the cell. When a message enters the LSTM network, the system would determine whether the message is useful or not. Only information that complies with algorithmic certification will remian, and misleading information will be forgotten through forgotten doors.

\begin{figure}[ht]
\centering
\includegraphics[scale=0.3]{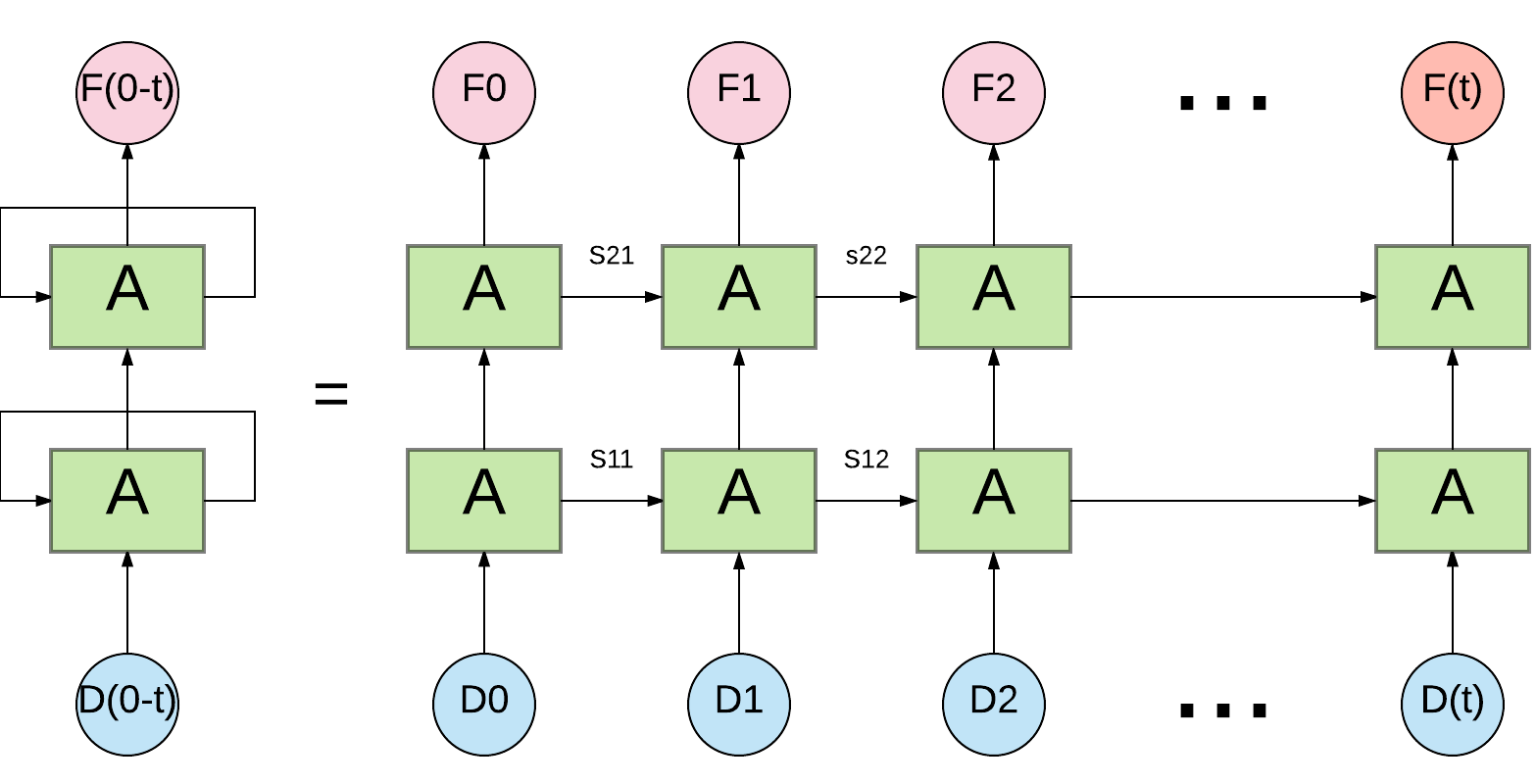}
\caption{Two layer LSTM recurrent network. Each green block is a LSTM block. Blue circle D(t) represent for sensor data recorded at time t, s is inner LSTM state, pink circle F is the output. Each LSTM cell receives the  input of current sensor data and state from previous cell (initial state is 0), and ouput a state  and a output.}
\label{fig:LSTM}
\end{figure}

\begin{figure*}[ht]
\centering
\includegraphics[scale=0.6]{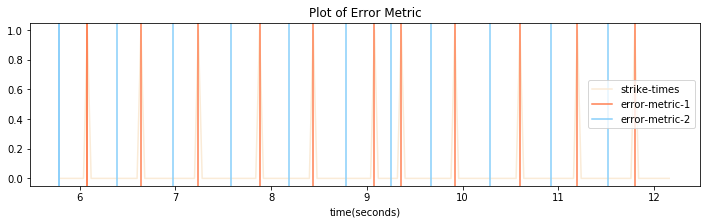}
\caption{An example of three error metrics. Assume above plot is a entire segement. Error metric 1 (read line) split intervals exactly at each heel strike time. Error metric 2 (blue line) split intervals at middle of every two continues heel strike times. Error metric 3 calculate undercount and overcount rate for the entire segment.}
\label{fig:error_metric}
\end{figure*}

We use TensorFlow to implement LSTM network. TensorFlow is an open-source software library that can be used for machine learning applications such as neural networks \cite{wiki:tensorflow}. It supports both CPU and GPU that can be imported as a python library. TensorFlow uses a data flow graph to represent computation in terms of the dependencies between individual operations \cite{tensorflow.org}. We first define the data flow graph and then create a session to run the graph. The saver class can save and restore variables to and from checkpoints by mapping variable names to tensor values.

As shown in picture \ref{fig:LSTM}, a two-layer RNN network with basic LSTM cell is built for training with squared difference loss function. 
As shown in picture \ref{fig:batch_sensor_data}, the input batch is a list of metrics. Each row of metrics is a list of sensor data at a time. The length of times is called timesteps.
For Example, we want to use previous timesteps ($=50$) record of sensor data to predict the result, each sensor data contains input number ($=6$) of values of rotation rate and user acceleration in 3-dimension. So the matrix size is timesteps multiply input number($50 \times 6$). All such metrics form the input list and is transformed to tensor as train data. Since the dataset is big and there are more than one hundred thousand elements in input list which would make the training process very slow, we shuffle and divide train data into the small batch (batch size = 256) and feed the batches to model one step by one step.

There are two ways of output. The first one  is to use previous timesteps data to predict only the last step signal value. The second way is to use previous timesteps data to predict corresponding timesteps step results. The first one cannot predict the first timesteps result, but it is more precise since all result is predicted by previous timesteps record.  Outputs that have enough previous records (timesteps previos record) are predicted output using first way. But the first timestep outputs of each segment don't have enough previous records, so they are predicted using the second way.

\begin{figure*}[ht]
\centering
\includegraphics[scale=0.5]{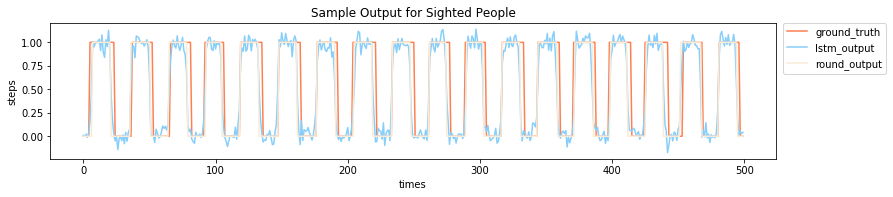}
\caption{Sample output of mixed data for sighted people. The origin output signal (blue line) swings near 0 or 1. The round signal (yellow line) is quite similar to the ground truth data (red line).}
\label{fig:output_sighted}
\end{figure*}

\subsection{Error Metrics}

The origin result signal is a float function around 0 to 1. Then we use 0.5 as the border to turn the signal into 0 or 1. And the predicted heel strike times is the signal change times. We can calculate the accuracy rate of the result, which is the proportion of correct numbers (1 is predicted to be 1 and 0 is predicted to be 0) and total numbers. The error rate of the step counting model was measured using three different metrics as shown in picture \ref{fig:error_metric}. The first two metric is same as error metric in \cite{flores2016weallwalk}.

The first metric split intervals at each ground-truth heel strike times, and counts the number of predicted steps within each time interval $[T_i, T_{i+1}]$.  Ifexactly one step is detected at each interval, the count of step is correct. If no step is detected during the interval, then one undercount event happens. If multiple steps are detected during the interval, then number of count-1 overcount events happen. The undercount and overcount rate is number of undercount and overcount event divided by the total count of ground-truth steps.

The second metric is similar to the first one,  but it  split intervals at middle of two continues ground-truth heel strike times, and counts the number of predicted steps within each time interval $[[\frac{T_{i-1}+T_i}{2}, \frac{T_i+T_{i+1}}{2}]$. The undercount and overcount is calculated same as first metric. As shown in picture \ref{fig:output_small_metric2} at end of the paper,  the metric can decrease error rate when one step is detected sightly earlier ($t_i<T_i$) and the next step is detected sighted later ($t_{i+1}<T_{i+1}$). For this case, interval $[T_i, T_{i+1}]$ has one undercount, and interval $[T_{i-1}, T_i]$ and $[T_{i+1}, T_{i+2}]$ have one overcount.

The third metric counts number of predicted steps during each segment and compare to the number of ground-truth steps during the same segment. If the ground-truth number is larger, then undercount events happen. If the predicted number is larger, then overcount events happen. The undercount and overcount rate is proportion between sum of undercount and overcount numbers over all segment and the total count of ground-truth steps. The undercount and overcount rate obtained by the error metric 3 metrics is always smaller than or equal to the corresponding values computed by the metric 1 and 2.

\begin{figure*}[ht]
\centering
\includegraphics[scale=0.5]{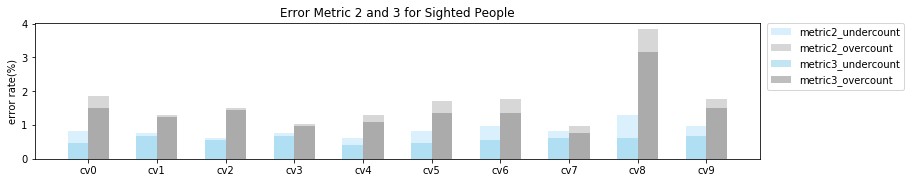}
\caption{Error metric 2 and 3 for 10 fold cross validation of mixed data for sighted people.The error rate of metric 2 is always higher than metric 3. The undercount error (blue blocks) is less than overcount error (grey blocks). Detailed values is shown in table \ref{label_metric23_sighted}.}
\label{fig:error_metric_23_na_10fold}
\end{figure*}

\begin{figure*}[ht]
\centering
\includegraphics[scale=0.5]{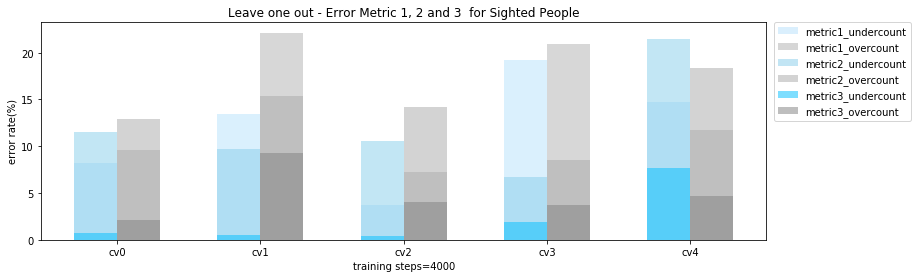}
\caption{Error metric 1, 2 and 3 for leave one out of sighted people.}
\label{fig:error_metric_123_na_5fold}
\end{figure*}

\section{Experience}

We preprocessed the WeAllWalk data to train LSTM model for the different communities using two ways of splitting train and test data. First way is simply mixed all data and split into train and test set. In this way, both train and test data contain records from all participant and all  segment. The second way is leave one person out, which means test set contains and only contains all records from one participant and train set contains all other records from remain participants.

The model is trained with different sensor data, timesteps, output number, hidden layer number, optimizer, learning rate and training steps. We first try different optimizers with various learning rate and find that AdamOptimizer with learning rate around 0.01 can make it convergence, and then fix the two parameters.

\subsection{Sighted people}
There are more than 120,000  valid records of sighted people, we split it into 100,000 train data and 25,000 test data. Each input data is a 3-dimensional tensor, with a shape of batch size $\times$ timesteps $\times$ number of sensor data.

For mixed data, we apply 10 folds cross-validation. A good result sample is shown in picture \ref{fig:output_sighted}. The output signal swings near 0 or 1. So it is reasonable to round the signal ($>0.5$ turn to 1, $<=0.5$ turn to 0). The round signal is similar to the ground truth data. However, not all result is as good as this sample, some bad samples are shown in end of the paper. Some result has a certain offset between ground truth data and predicted output, some result function shakes more fiercely so that not all values are close to 0 and 1. There tend to be more strikes if the output value shakes around 0.5. So in the experience the overcount rate is always higher than the overcount rate.

\begin{figure}[ht]
\centering
\includegraphics[scale=0.4]{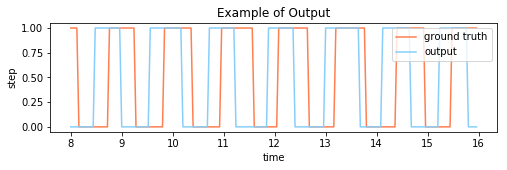}
\caption{Sample output of offset. In this case, error metric 1 works better than error metric 2.}
\label{fig:output_ex_offset}
\end{figure}

Since there are 6 error values, it difficult to determine a weighted format over each error rate and tell which model is better. All ten results are shown in picture \ref{fig:error_metric_23_na_10fold}.
The average undercount rate of metric 3 is around 0.6\%, and the average overcount rate of metric 3 is around 1.4\%.  
The average undercount rate of metric 2 is around 0.9\%, and the average overcount rate of metric 2 is around 1.7\%.  
The value of error rate is shown in table \ref{label_metric23_sighted}. The error rate of metric 1 is much higher than error rate of metric 2 and 3, the undercount rate of metric 1  is similar to overcount rate of metric 1 (around 20\%) as shown in picture \ref{fig:error_metric_1_na_10fold}. The reason that caused such situation is described in picture \ref{fig:output_small_metric2}. So the metric 1 error rate of undercount and overcount is similar since when a undercount happened it is likely to cause a overcount.

\begin{figure}[ht]
\centering
\includegraphics[scale=0.55]{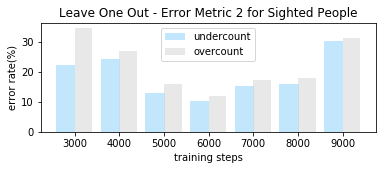}
\caption{Error metric 1 and 3 for 10 fold cross validation of sighted people.The error rate of metric 1 is always higher than metric 3 with regrad to training steps.}
\label{fig:error_metric_2_na_step}
\end{figure}

\begin{figure*}[ht]
\centering
\includegraphics[scale=0.5]{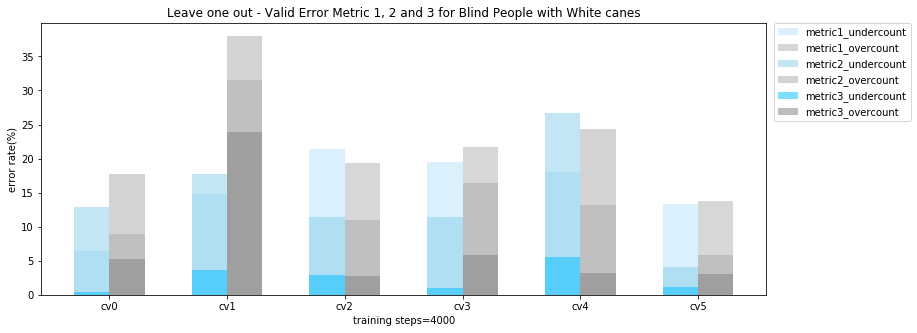}
\caption{Leave one out - Valid Error for Blind People with a long cane}
\label{fig:error_metric_wc_10fold_valid4000}
\end{figure*}

\begin{figure*}[ht]
\centering
\includegraphics[scale=0.5]{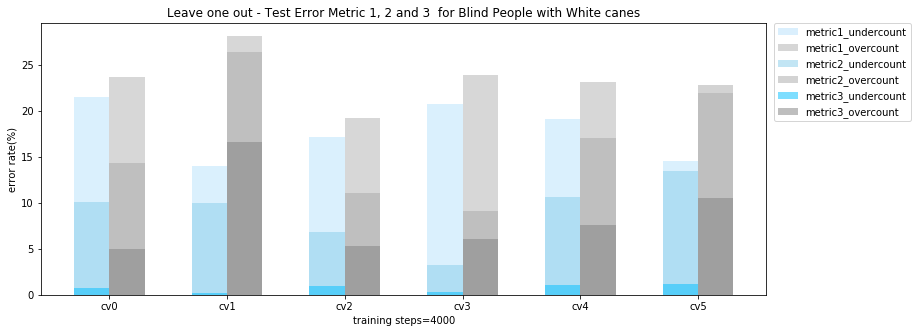}
\caption{Leave one out - Test Error for Blind People with a long cane}
\label{fig:error_metric_wc_10fold_test4000}
\end{figure*}

For leave one person out method, the cross validation result is shown in picture \ref{fig:error_metric_123_na_5fold}.
The average undercount rate of metric 3 is around 2.5\%. The average overcount rate of metric 3 is around 7\%.
 In some case, metric 2 has bad performance, this is because there is an offset around half of each interval as shown in the picture  \ref{fig:error_metric_2_na_step}. 
 In the experience with different parameters of LSTM model,  some vaildation sets (cv3 perform best in most of experience) have overcount and undercount rate less than 0.1\% of metric 3, while less than 1\% for metric 1 or 2. But for other validation set, the overcount rate is extremely high likely 30\% to 60\%. Each vaildation set has better performance with models of different parameters. It may because different people have different gaits. So I choose the model where each validation sets have average performance. 

When the train and test data both contains the sensor data from certain people,  the predicted signal is quite accurate, and the error (metric 2 and 3) of step counting is small. However, for leave one out method, there would be offsets for both train and test data, and the accuracy of predicted signal is low. But the error 1 could be small if the offset is similar for each heel strikes as shown in picture \ref{fig:output_ex_offset}. 

\subsection{Blind People with a long cane}
There are more than 120,000 records from blind people with a long cane, participants 1, 2, 3, 5, 6, 7, 8 have sensor data for six paths. So we use participant 8 as test data, and records from remain six people as train and validation data for 6 fold cross validation for leave one out. The valid and test result is shown in picture \ref{fig:error_metric_wc_10fold_valid4000} and \ref{fig:error_metric_wc_10fold_test4000}. All error rate is higher than mix data. The validation set has good performance don't means that the test data also have good performance.
The training data only contains six different people, which is not a big number. If we have annotated data from more people, LSTM model may detect common features of blind people and provide a better result.

\begin{figure*}[ht]
\centering
\includegraphics[scale=0.5]{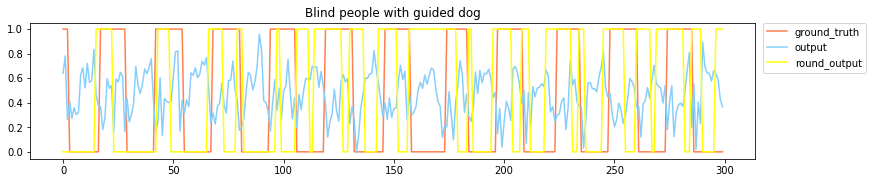}
\caption{Sample output of blind people with guide dog.}
\label{fig:blind_gd}
\end{figure*}

\subsection{Blind walkers using a guide dog}

There are only 3 people walked with the guide dog, each has 30,000 data. So we regard two people as train data, and the left one as test data. The result is bad as shown in \ref{fig:blind_gd}. Although mixed data shows good performance around 5\% for metric 2 and 2\% for metric 3, it is not likely to predict the strike heel times of a person from only one different person.

\begin{table}[]
\centering
\caption{Error of blind people with guide dog}
\label{my-label}
\begin{tabular}{llll}
\hline
error(\%)                &            & valid & test  \\ \hline
metric1                  & undercount & 23.90 & 10.18 \\
                         & overcount  & 26.14 & 25.75 \\ \hline
\multirow{2}{*}{metric2} & undercount & 22.56 & 12.27 \\
                         & overcount  & 29.03 & 19.49 \\ \hline
\multirow{2}{*}{metric3} & undercount & 4.52  & 0.53  \\
                         & overcount  & 10.98 & 17.75 \\ \hline
\end{tabular}
\end{table}

\section{Conclusion and Future Work}

We train an LSTM model using annotated smartphone sensor data from the WeAllWalk dataset, to predict output signal. In the dataset, blind volunteers using a long cane or a guide dog, sighted volunteers have different features of gaits, so we separately build the model and calculate the error rate of three metrics. We also apply leave one person out pedometers to test the accuracy of our mode using three error metrics, whihc splitting intervals differently to estimate the overcount and undercount of steps.  The model achieved 1\% overcount and undercount rate for mixed training data and 5\% for Leave-One-Out training modality.

As shown in picture \ref{fig:overcount} and picture \ref{fig:undercount}, simply round the origin output signal may cause the overcount or undercount. So we can use a better way to turn the signal into binary signal. In this paper, we only use the data of straight segments, removing turn segment and other features. We could consider about turn segments and feature motions. We now only use rotation rate and user acceleration as inout sensor data, we can also try more sensor data in future.

\begin{figure}[ht]
\centering
\includegraphics[scale=0.4]{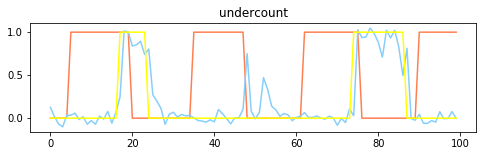}
\caption{Sample of undercount.}
\label{fig:undercount}
\end{figure}

\begin{figure}[ht]
\centering
\includegraphics[scale=0.4]{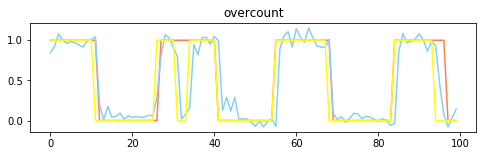}
\caption{Sample of overcount.}
\label{fig:overcount}
\end{figure}

\begin{figure*}[ht]
\centering
\includegraphics[scale=0.5]{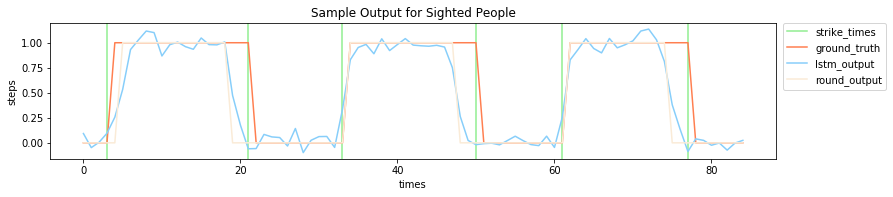}
\caption{Bad sample for error metric 1. The green line splits the intervals of error metric 1. There are random offset between actual strikes times and predicted strike times. In this case, the first, third and fifth interval has one overcount, second and fourth interval has one undercount.}
\label{fig:output_small_metric2}
\end{figure*}

\begin{figure*}[ht]
\centering
\includegraphics[scale=0.5]{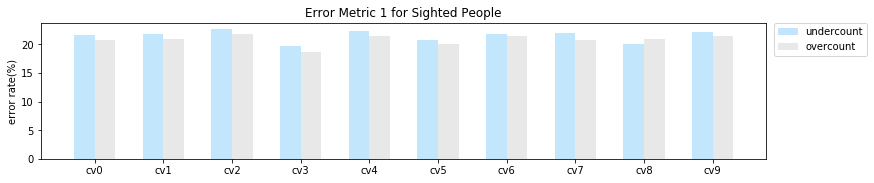}
\caption{Error metric 1 for 10 fold cross validation for sighted people. The error rate is much higher than error rate for metric 2 and 3 as shown in picture. The reason that caused such situation is described in picture.}
\label{fig:error_metric_1_na_10fold}
\end{figure*}

\begin{figure*}[ht]
\centering
\includegraphics[scale=0.5]{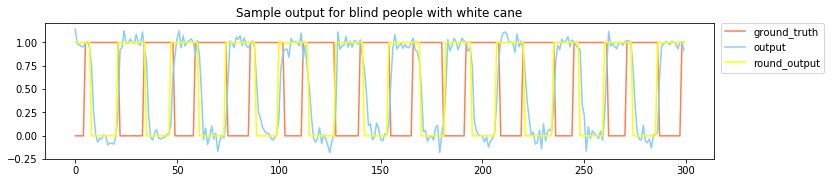}
\caption{Sample output of blind people with long cane.}
\label{fig:output_wc_1}
\end{figure*}

\begin{figure*}[ht]
\centering
\includegraphics[scale=0.5]{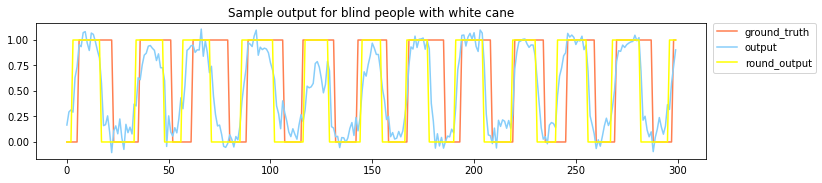}
\caption{Sample output of blind people with long cane.}
\label{fig:output_wc_2}
\end{figure*}

\begin{table*}[]
\centering
\caption{Error metric 2 and 3 for 10 fold cross validation of sighted people}
\label{label_metric23_sighted}
\begin{tabular}{llllllllllll}
\hline
error(\%)                &            & cv0   & cv1   & cv2   & cv3   & cv4   & cv5   & cv6   & cv7   & cv8   & cv9  \\ \hline
metric1                  & undercount & 0.82  & 0.75  & 0.62  & 0.75  & 0.62  & 0.82  & 0.96  & 0.82  & 1.3   & 0.96 \\
                         & overcount  & 1.85  & 1.3   & 1.5   & 1.03  & 1.3   & 1.71  & 1.78  & 0.96  & 3.83  & 1.78 \\ \hline
\multirow{2}{*}{metric2} & undercount & 0.48  & 0.68  & 0.55  & 0.68  & 0.41  & 0.48  & 0.55  & 0.62  & 0.62  & 0.68 \\
                         & overcount  & 1.5   & 1.23  & 1.44  & 0.96  & 1.09  & 1.37  & 1.37  & 0.75  & 3.15  & 1.5  \\ \hline
\end{tabular}
\end{table*}

\begin{table*}[]
\centering
\caption{Error mertric 1, 2 and 3 for blind walkers using a long cane}
\label{my-label}
\hspace*{-1.5cm}
\begin{tabular}{|l|l|l|l|l|l|l|l|l|l|l|l|l|l|}
\hline
\multicolumn{2}{|l|}{\multirow{2}{*}{error(\%)}} & \multicolumn{2}{l|}{cv0} & \multicolumn{2}{l|}{cv1} & \multicolumn{2}{l|}{cv2} & \multicolumn{2}{l|}{cv3} & \multicolumn{2}{l|}{cv4} & \multicolumn{2}{l|}{cv5} \\ \cline{3-14} 
\multicolumn{2}{|l|}{}                           & valid       & test       & valid       & test       & valid       & test       & valid       & test       & valid       & test       & valid       & test       \\ \hline
\multirow{2}{*}{m1}       & undercount      & 6.44        & 21.49      & 14.83       & 14.03      & 21.45       & 17.1       & 19.5        & 20.67      & 18.05       & 19.03      & 13.35       & 14.56      \\ \cline{2-14} 
                               & overcount       & 8.87        & 23.63      & 31.5        & 28.1       & 19.32       & 19.2       & 21.69       & 23.91      & 13.17       & 23.09      & 13.7        & 21.94      \\ \hline
\multirow{2}{*}{m2}       & undercount      & 12.93       & 10.09      & 17.79       & 9.97       & 11.36       & 6.77       & 11.49       & 3.28       & 26.69       & 10.62      & 4.09        & 13.45      \\ \cline{2-14} 
                               & overcount       & 17.81       & 14.36      & 38.01       & 26.33      & 10.97       & 11.07      & 16.37       & 9.06       & 24.35       & 17.06      & 5.9         & 22.76      \\ \hline
\multirow{2}{*}{m3}       & undercount      & 0.43        & 0.7        & 3.63        & 0.21       & 2.94        & 0.94       & 1.03        & 0.29       & 5.59        & 1.11       & 1.2         & 1.19       \\ \cline{2-14} 
                               & overcount       & 5.32        & 4.96       & 23.85       & 16.57      & 2.7         & 5.25       & 5.9         & 6.07       & 3.25        & 7.55       & 3.01        & 10.5       \\ \hline
\end{tabular}
\end{table*}

\nocite{*}

\bibliographystyle{acl_natbib}
\bibliography{acl2015}

\end{document}